\documentclass[10pt,twocolumn,letterpaper]{article}

\usepackage{relsize}
\usepackage{wacv}
\usepackage{times}
\usepackage{epsfig}
\usepackage{graphicx}
\usepackage{amsmath}
\usepackage{amssymb}
\usepackage{blindtext}
\usepackage{graphicx}
\usepackage{caption}
\usepackage{subcaption}
\usepackage{enumitem}
\usepackage{booktabs}
\usepackage{array}
\usepackage{algorithm} 
\usepackage{algorithmic}  
\usepackage[algo2e]{algorithm2e} 
\usepackage{amsmath}
\usepackage{amsfonts}
\usepackage{amssymb}
\usepackage[pagebackref=true,breaklinks=true,letterpaper=true,colorlinks,bookmarks=false]{hyperref}
\usepackage{cleveref}

\wacvfinalcopy 

\def\wacvPaperID{****} 

\ifwacvfinal\fi
\begin{document}

\title{Deep Image Blending}


\author{Lingzhi Zhang\qquad Tarmily Wen\qquad Jianbo Shi \\University of Pennsylvania\\{\tt\small \{zlz, jshi\}@seas.upenn.edu, weng@sas.upenn.edu}}


\twocolumn[{%
\renewcommand\twocolumn[1][]{#1}%
\maketitle
\begin{center}
    \centering
    \includegraphics[width=\textwidth]{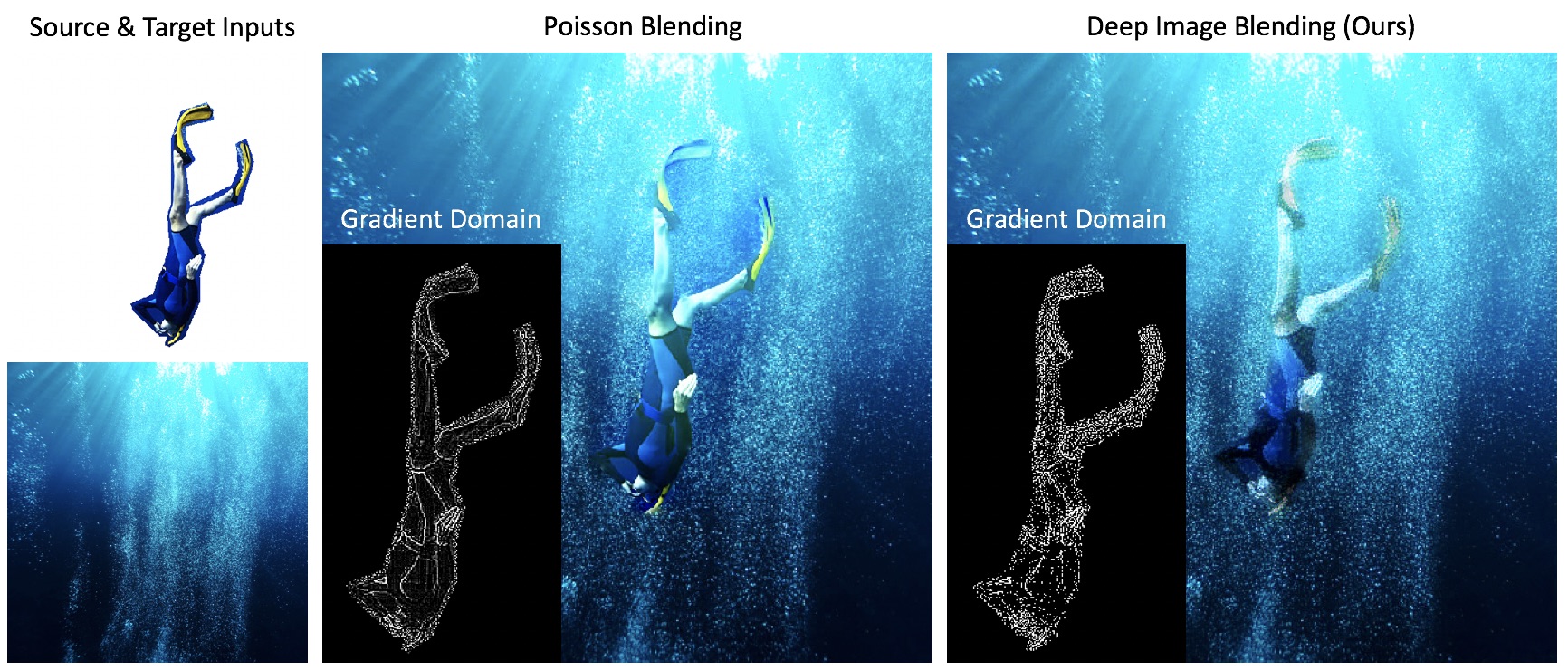}
    \captionof{figure}{Given a coarsely cropped object in a source image, a target image and a blending location, our algorithm can blend the selected object onto a target image with seamless boundary and consistent style with respect to the target image. 
    }
    \label{fig:header}
\end{center}%
}]

\begin{abstract}
\vspace{-5 pt}

Image composition is an important operation to create visual content. Among image composition tasks, image blending aims to seamlessly blend an object from a source image onto a target image with lightly mask adjustment. A popular approach is Poisson image blending \cite{poisson_editing}, which enforces the gradient domain smoothness in the composite image. However, this approach only considers the boundary pixels of target image, and thus can not adapt to texture of target image. In addition, the colors of the target image often seep through the original source object too much causing a significant loss of content of the source object. We propose a Poisson blending loss that achieves the same purpose of Poisson image blending. In addition, we jointly optimize the proposed Poisson blending loss as well as the style and content loss computed from a deep network, and reconstruct the blending region by iteratively updating the pixels using the L-BFGS solver. In the blending image, we not only smooth out gradient domain of the blending boundary but also add consistent texture into the blending region. User studies show that our method outperforms strong baselines as well as state-of-the-art approaches when placing objects onto both paintings and real-world images. Code is available at: \href{https://github.com/owenzlz/Deep_Image_Blending}{https://github.com/owenzlz/DeepImageBlending}
\end{abstract}

\vspace{-10 pt}
\section{Introduction}
\vspace{-5 pt}

Image blending is a method for image composition. It generally refers to cropping a certain region of a source image (usually an object) and placing it onto the target image at a specified location, where the goal is to make the composite image look as natural as possible. The challenge of this task is that the cropped region may not be precisely delineated. Therefore, the blending process needs to not only adjust the appearance of the cropped object to be compatible with the new background but also make the cropping boundary appear seamless.

The current most popular method for image blending is Poisson image editing \cite{poisson_editing}. The idea is to reconstruct pixels in the blending region such that the blending boundary has smooth pixel transition or small gradients with respect to the boundary pixels in the target image. However, this method is difficult to combine with other reconstruction objectives because of its closed-form matrix solution. A recent work \cite{gpgan} combines the closed-form solution of Poisson equation with GAN loss to synthesize realistic blending images. However, this method requires a source region, a target image, and a corresponding well-blended image as training examples for supervised learning. Since such data is extremely rare and expensive to label, the generalization and application domain of this method is limited. Closely related to image blending is image harmonization, but the foreground object must be precisely delineated and thus the goal is to only adjust the illumination, color, and texture of the foreground to make it compatible with the new background. 

In this work, we propose a novel two-stage blending algorithm. The algorithm first generates a seamless boundary for the source region, and then further refines the region with similar styles and textures with respect to the target image. In this algorithm, we propose a differentiable loss that enforces the equivalent purpose of the original objective of Poisson equation, and it can be easily combined with other reconstruction objective functions. Our algorithm works well for not only real-world target images but also stylized paintings by utilizing content and style loss \cite{style1} from deep features. In addition, our algorithm solves the reconstruction of image blending using only a single source image, a coarse mask, and a target image. Since our algorithm does not rely on any training data, it can generalize to any source and target images. Finally, we show the uniqueness and effectiveness of our algorithm compared to the state-of-the-arts methods through various testing cases.

\vspace{-5 pt}
\section{Related Work} 

\subsection{Image Blending}

Image blending refers to cropping a certain region of a source image (usually an object) and placing it onto the target image at a specified location, where the goal is to make the composite image look as natural as possible. In contrast with image harmonization, an important characteristic of image blending is that it does not need precise object delineation for the blending mask. The default way of doing this task is to directly copy pixels from source image and paste them onto the target image, but this would generate obvious artifacts because of the abrupt intensity change in the compositing boundaries. 

An early work, alpha blending \cite{alpha_blending}, is the process of layering multiple images, with the alpha value for a pixel in a given layer indicating what fraction of the colors from lower layers are seen through the color at the given level. Although alpha blending performs much better than directly copy-and-pasting, it produces a ghost effect in the composite image as the contents in both source and target images exist in the same region. 

Alternatively, the most popular image blending technique aims to inform gradient domain smoothness \cite{poisson_editing,gb1,gb2,gb3,gb4,gb5,gb6,gb7}. The motivation of gradient domain blending is that the human visual system is very sensitive to the regions with abrupt intensity change, such as edges, and thus we want to produce an image with smooth transition over the blending boundary. The earliest work \cite{poisson_editing} proposes to reconstruct the pixels of the blending region in the target image by enforcing the gradient domain consistency with respect to the source image, where the gradient of the blending region is computed and propagated from the boundary pixels in the target image. With such gradient domain consistency, the blended image will have smooth transitions over the composite boundary even though the object mask are not precisely delineated. Our work is partially inspired by Poisson image editing \cite{poisson_editing}, which will be further described in section 3. 

A recent approach GP-GAN \cite{gpgan} has leveraged the closed-form solution of the Gaussian-Poison Equation \cite{Gaussian_Poisson} and Wasserstein Generative Adversarial Network (WGAN) \cite{wgan} to produce photo-realistic blending results. However, this approach relies on supervised training, which requires paired data of a source image, target image, and corresponding well-blended image as ground-truth, so the generalization is difficult.

\subsection{Other Image Editing Techniques}

Other common image editing tools includes image denoising, image super-resolution, image inpainting, image harmonization, style transfer and so on. In recent years, Generative Adversarial Networks (GANs) \cite{gan} have been extensively applied to these tasks and produced very exciting results.

In super resolution \cite{SR1, sr2, SR3_style2, sr4, sr5, sr6, sr7}, deep models learn the image texture prior to upsample a low-resolution image into high-resolution version. In image inpainting \cite{inpaint1, inpaint2, inpaint3, inpaint4, inpaint5, inpaint6, inpaint7}, the network aims to fill the missing pixels in an image with the learned semantic information as well as real image statistics.

\begin{figure*}[!t]
\centering
\includegraphics[width=\textwidth]{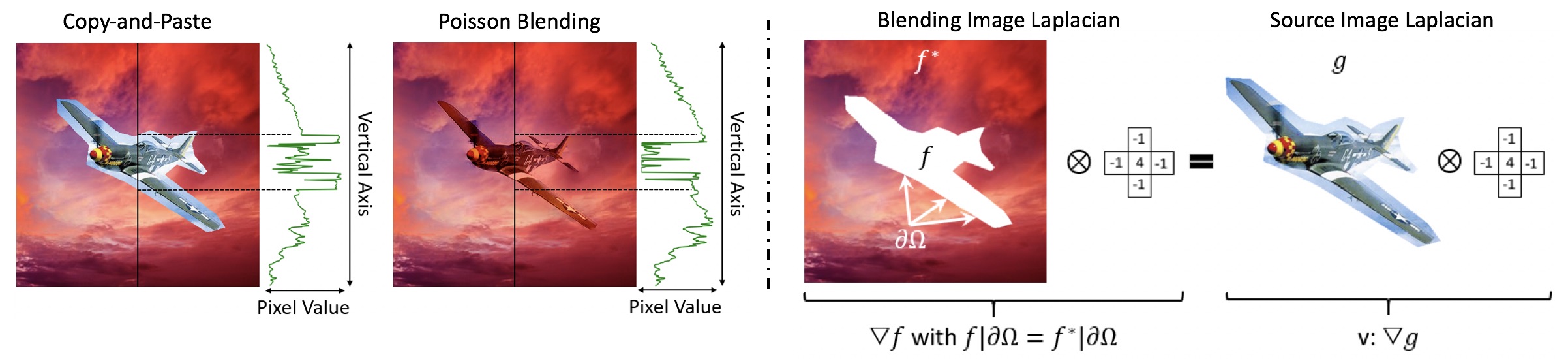}
\caption{The two images on the left show the intensity change in a "copy-and-paste" image and a "Poisson Blended" image. The images on the right demonstrate the essential idea of poisson image editing, where the goal is to enforce the Laplacian gradient domain consistency between the source image and the blended image with a boundary constraint from target image. Specifically, we use a laplacian filter to compute the second order gradient of images. }
\label{fig:poisson}
\vspace{-10 pt}
\end{figure*}

Closely related to our task is image harmonization, which extracts the foreground region in one image and combines it with the background of another image while adjusting the appearance of the foreground region to make it compatible with the new background. Early works \cite{harm1, harm2} use the global color statistics of the background image to adjust the appearance of the foreground image. Recently, \cite{harm6} propose an automatic data acquisition approach and learn an end-to-end deep network to capture both the context and semantic information of the composite image during harmonization. Closest to our work, \cite{harm7} propose a deep painterly harmonization technique to composite a real-world object onto a stylized paintings by adjusting the parameters of the transfer depending on the painting.

Another closely related task is style transfer \cite{style1, SR3_style2, style3, style4, style5, histogram_loss}, which aims to transform the style of an image into the style of another image. \cite{style1} first propose to transform a content image into the style of another image by jointly optimizing the transformed image's similarity of deep features with respect to the content image and similarity of gram matrices of deep features with respect to the style image.




\vspace{-5 pt}
\section{Background -  Poisson Image Editing}

Directly copying a foreground object from source image and pasting it onto a target image can produce big intensity changes at the boundary, which creates obvious artifacts to human eyes. Therefore, the motivation of poisson image blending is to smooth the abrupt intensity change in the blending boundary in order to reduce artifacts. In Fig.(\ref{fig:poisson}), the left image shows the abrupt intensity change in the composite boundary in the copy-and-paste image and a smooth boundary transition using poisson blending \cite{poisson_editing}.


In the original work \cite{poisson_editing}, poisson image blending is formulated as an image interpolation problem using a guidance vector field. 

\begin{equation}
    \min_{f} \iint_{\Omega} |\triangledown f - \textbf{v}|^2  \mbox{ with } f|_{\partial \Omega} = f^*|_{\partial \Omega}
    \label{Eq:poisson}
\end{equation}

where $\nabla = [\frac{\partial.}{\partial x},\frac{\partial.}{\partial y}]$ is the gradient operator, $f$ is the function of the blending image, $f^*$ is the function of the target image, \textbf{v} is the vector field, $\Omega$ is the blending region and $\partial \Omega$ is the boundary of the blending region. In this case, the guidance field \textbf{v} is the gradient field taken directly from source image $g$. 

\begin{equation}
    \textbf{v} = \triangledown g
\end{equation}

We solve this minimization problem with boundary condition for each color channel independently to obtain the RGB image. 


For images, the problem can be discretized using the underlying discrete pixel grid to obtain a quadratic optimization problem. 
\begin{equation}
\underset{f|_{\Omega}}{\mathrm{min}} \sum_{ \left \langle p,q \bigcap \Omega \neq \emptyset \right \rangle}(f_p - f_q - v_{pq})^2 \text{, with } f_p=f_p^* \text{ for all } p \in \partial \Omega
\end{equation}
where $N_p$ is the set of 4-connected neighbors for pixel $p$, $\left \langle p,q \right \rangle$ denote a pixel pair such that $q \in N_p$, $f_p$ is the value of $f$ at $p$ and $v_{pq}=g_p-g_q$  for all $\left \langle p,q \right \rangle$.

\begin{figure*}[!t]
\centering
\includegraphics[width=\textwidth]{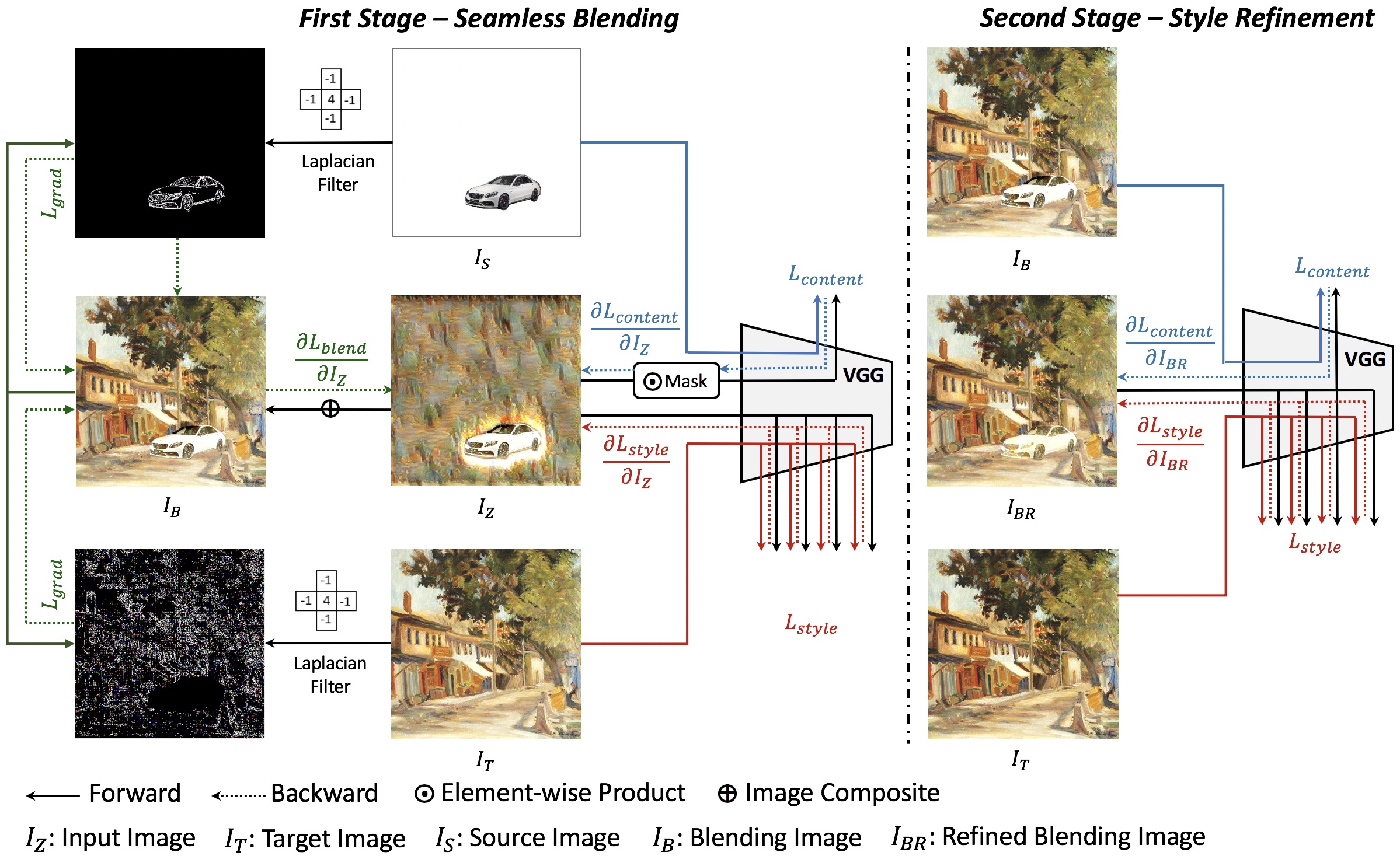}
\caption{This figure shows our two-stage blending algorithm. In the first stage, an input image $I_Z$ is randomly initialized and directly gets updated with respect to a gradient loss,  content loss, and style loss. The gradient loss enforces the gradient of blending region to be the same as the gradient of the source object, the content loss enforces the semantic similarity between the blending region and the source object, and the style loss enforces the textural similarity between the blending region and the target image. In the second stage, the blending image from first stage is considered as an input image, and is further optimized with respect to the blending image and target image in terms of content and style respectively. }
\label{fig:architecture}
\vspace{-10 pt}
\end{figure*}

For discrete system, the solution can be converted into the following simultaneous linear equations. For $p \in \Omega$, the linear equations are as follows: 

\begin{equation}
  |N_p|f_p - \sum_{q \in N_p \bigcap \Omega} f_q= \sum_{q \in N_p \bigcap \partial \Omega} f_q^*+\sum_{q \in N_p} v_{pq}
\label{equation4}
\end{equation}

For pixels $p$ interior to $\Omega$, there are no boundary pixels from the target image. Thus, the equation becomes the following: 

\begin{equation}
|N_p|f_p - \sum_{q \in N_p} f_q=\sum_{q \in N_p} v_{pq}
\end{equation}

We need to solve for $f_p$ from the given set of simultaneous linear equations. Since Eq.(\ref{equation4}) form a sparse symmetric positive-definite system, two classical iterative solvers Gauss-Seidel and V-cycle multigrid\cite{poisson_editing} are used to solve the linear system in the early works.

\section{Methods}

Our algorithm is a two-stage process, as shown in Fig.(\ref{fig:architecture}). In the first stage, a preliminary blended image is synthesized using the proposed Poisson gradient loss, style loss, and content loss. In the second stage, the preliminary blended image is further transformed to have more a similar style to match the target image. Here, we denote $I_S$ as source image, $I_T$ as target image, $I_B$ as blending image, $I_{BR}$ as refined blending image, and $M$ as mask. Here, we assume the source image $I_S$ has already been cropped out using the coarse mask $M$. The size of $I_S$ and $I_T$ may or may not be the same, but they can be easily aligned using the user-provided offsets. For simplicity, we consider $I_S$ and $M$ are already aligned with $I_T$ and thus have the same dimension in the following discussion. We further define an input image as $I_Z$, which represents the reconstructed pixels. During training, the joint loss back-propagates to $I_Z$ in stage one or $I_{BR}$ in stage two, and thus the optimization process essentially adjusts pixel values in $I_Z$ or $I_{BR}$. 

\begin{figure*}[!t]
\centering
\includegraphics[width=\textwidth]{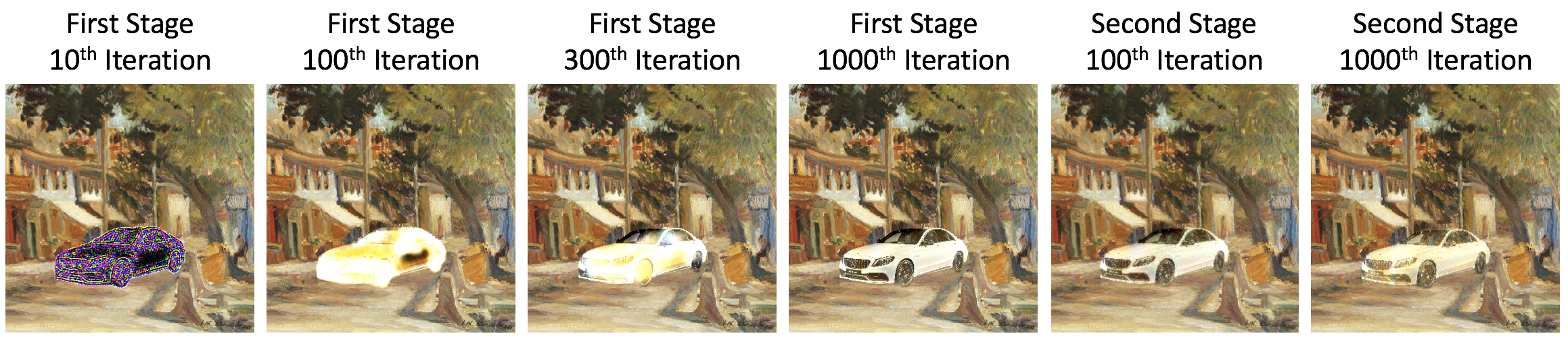}
\caption{This is a demonstrate of the blending pixel reconstruction process at different iterations in stage one and stage two. }
\label{fig:recon_process}
\vspace{-10 pt}
\end{figure*}

\subsection{Poisson Gradient Loss}

As we discuss in Section 3, Poisson image equation (\ref{Eq:poisson}) proposed to reconstruct the pixels of the blending region by enforcing the gradient domain consistency between the blending image and the source image. In the meantime, the gradient of the blending image is initially computed from the boundary pixels of the target image and propagated toward the inside. Such consistency produce seamless boundary of the blending region, but it is solved using a well-designed matrix operation and is difficult to combine with other constraints for pixel reconstruction. Thus, we propose to convert this gradient domain constraint into a differentiable loss function as follows, 

\begin{equation}
    \mathcal{L}_{grad} = \frac{1}{2 H W} \sum_{m=1}^{H} \sum_{n=1}^{W} [\triangledown f(I_B) - (\triangledown f(I_S) + \triangledown f(I_T))]^2_{mn}
    \label{Eq:L_grad}
\end{equation}

In Eq.(\ref{Eq:L_grad}), $\triangledown$ represents the Laplacian gradient operator, and $H$ and $W$ are the width and height of image. The blending image is defined as $I_B = I_Z \odot M + I_T \odot (1-M)$. This loss function is equivalent to the Poisson Eq.(\ref{Eq:poisson}). First, the reconstructed pixels of $I_Z$ is directly combined with $I_T$ to construct $I_B$, and then the Laplacian filter is operated on the whole $I_B$, which takes the boundary pixels of $I_T$ into account. This part satisfies the boundary constraint in Poisson equation. Second, we directly minimize the difference between the gradient of $I_B$ and the addition of gradients of $I_S$ and $I_T$. Since the gradient of $I_T$ is exactly the same as the gradient outside blending region in $I_B$, the loss is essentially computed within the blending region. The second part satisfies the gradient domain constraint in Eq.(\ref{Eq:poisson}). An alternative way to implement this loss is to crop out the gradient of the blending region in $I_B$ using $M$ and compare it with $\triangledown I_S$. We think these two ways of implementation make little difference. 

\subsection{Style and Content Loss}

In the original work \cite{style1}, Gatys et al proposed to transform the style of a source image using a style loss while preserving the content of the source using a content loss. In the first stage, the content and style losses are defined as follows,

\begin{equation}
    \mathcal{L}_{cont} = \sum_{l=1}^{L} \frac{\alpha_l}{2 N_l M_l} \sum_{i=1}^{N_l} \sum_{k=1}^{M_l} (F_l[I_Z]\ \odot M - F_l[I_S])^2_{ik}
    \label{Eq:L_cont_first}
\end{equation}

\begin{equation}
    \mathcal{L}_{style} = \sum_{l=1}^{L} \frac{\beta_l}{2 N_l^2} \sum_{i=1}^{N_l} \sum_{j=1}^{N_l} (G_l[I_Z] - G_l[I_T])^2_{ij}
    \label{Eq:L_style_first}
\end{equation}

where $\odot$ is the element-wise product, $L$ is the number of convolutional layers, $N_l$ is the number of channels in activation, $M_l$ is the number of flattened activation values in each channel. $F_l[\cdot] \in \mathbb{R}^{N_l \times M_l}$ is an activation matrix computed from a deep network $F$ at the $l^{th}$ layer. $G_l[\cdot] = F_l[\cdot] F_l[\cdot]^T \in \mathbb{R}^{N_l \times N_l}$ denotes the Gram matrix of the corresponding activation matrix at the $l^{th}$ layer. Intuitively, the Gram matrix captures the similarity relation between all pairs of channel features, which encodes the image style or texture and zero information about spatial structure. Finally, $\alpha_l$ and $\beta_l$ are the weights that control the influence of each layer when computing the content and style loss. 

In the second stage, the inputs to the content and style losses are different, which are defined as follows,

\begin{equation}
    \mathcal{L}_{cont} = \sum_{l=1}^{L} \frac{\alpha_l}{2 N_l M_l} \sum_{i=1}^{N_l} \sum_{k=1}^{M_l} (F_l[I_{BR}] - F_l[I_B])^2_{ik}
    \label{Eq:L_cont_second}
\end{equation}

\begin{equation}
    \mathcal{L}_{style} = \sum_{l=1}^{L} \frac{\beta_l}{2 N_l^2} \sum_{i=1}^{N_l} \sum_{j=1}^{N_l} (G_l[I_{BR}] - G_l[I_T])^2_{ij}
    \label{Eq:L_style_second}
\end{equation}

where $I_{BR}$ is the refined blending image, which is optimized with respect to $I_B$ in terms of content and $I_T$ in terms of style.

\begin{figure*}[!t]
\centering
\includegraphics[width=\textwidth]{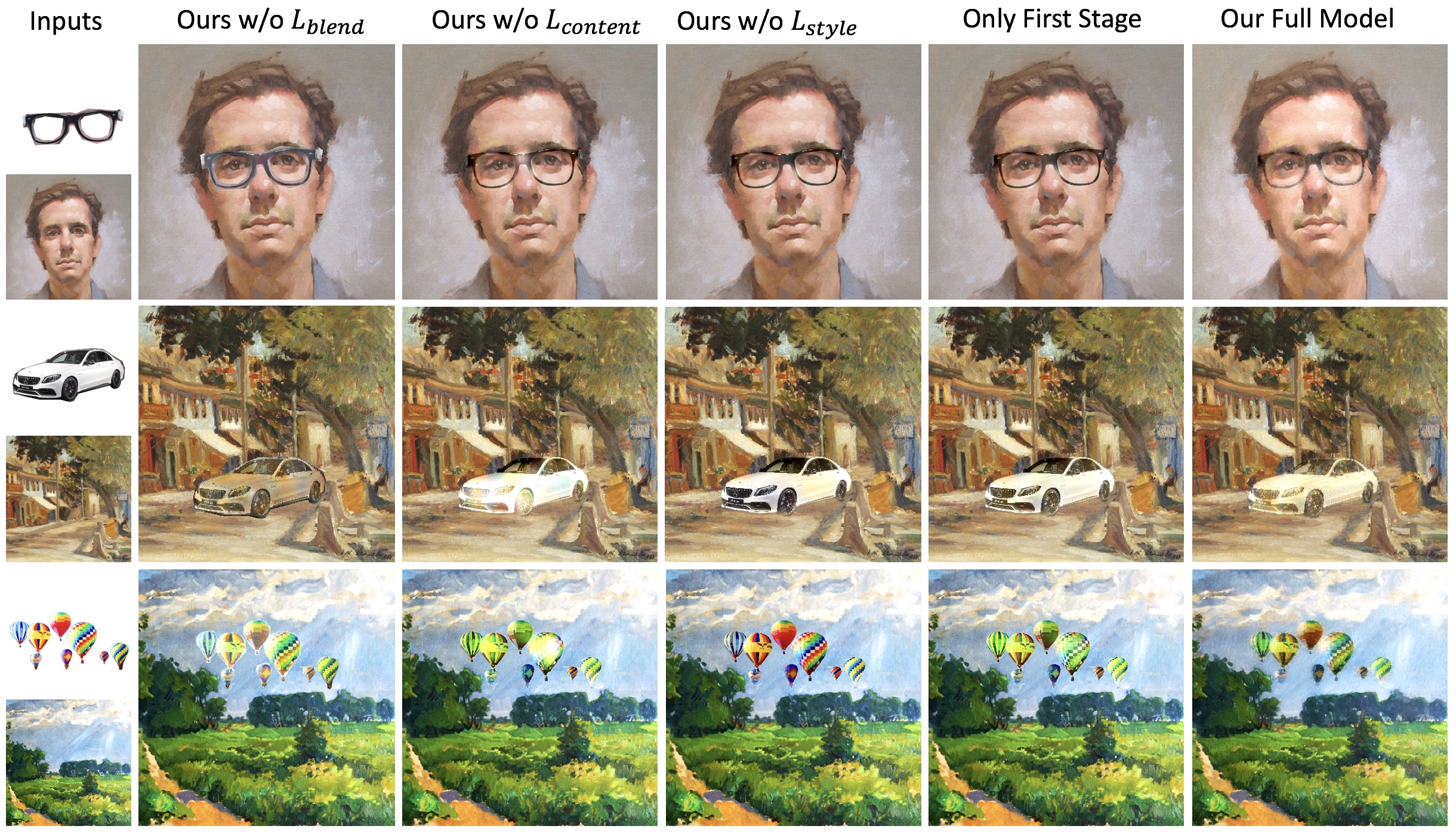}
\caption{Ablation study of different loss functions as well as single-stage versus two-stage. }
\label{fig:ablation_study}
\vspace{-10 pt}
\end{figure*}

\subsection{Regularization Loss}

To stabilize the style transformation of blended region and encourage spatial smoothness, we further add a histogram loss proposed by \cite{histogram_loss} and a total variation loss proposed by \cite{tv_loss} to regularize the generated image. 

The histogram loss from  \cite{histogram_loss} performs a histogram matching on each corresponding feature map at each output layer between the target image and the blended output. In this case we do it each iteration. Let $F_l(I_B)$ be the activation output for each layer of the blended image and $R_l(I_B)$ be the histogram matched activation between the blended output and the target image. This loss serves to stabilize the style transfer by matching the marginal distributions for each filter in each layer of the blended image with the marginal distribution of the target image. The activations we used for the histogram loss is the same activations as the style loss.

\newcommand{\norm}[1]{\left\lVert#1\right\rVert}

\begin{equation}
    \mathcal{L}_{hist} = \sum_{l=1}^{L} \gamma_l \norm{ F_l(I_B) - R_l(I_B) }\textsubscript{\raisebox{-1pt}{\rlap{F}}}\textsuperscript{\raisebox{1pt}{2}}
    \label{Eq:L_hist}
\end{equation}

The total variation (tv) loss is used to remove the unwanted details while preserving the import information in the image. The loss objective is shown below. 

\begin{equation}
    \mathcal{L}_{tv} = \sum_{m=1}^{H} \sum_{n=1}^{W} |I_{m+1, n}  - I_{m,n}| + |I_{m, n+1} - I_{m,n}|
    \label{Eq:L_tv}
\end{equation}

\subsection{Two-Stage Algorithm}

In our algorithm, the first stage aims to seamlessly blend the object onto the background, and the second stage aims to further refine the texture and style of the blending region. The input to the first stage is a 2D random noise, while the input to the second stage is the final blending image from the first stage. We use the VGG-16\cite{vgg} network pretrained on ImageNet\cite{imagenet} to extract features for computing style and content losses. Regarding the gradient domain, we use Laplacian filter to compute the second order gradient of images to compute the gradient blending loss.

\begin{algorithm}
\SetAlgoLined
 \textbf{Input}: source image $I_S$, blending mask $M$, target image $I_T$\par
 max iteration $T$, loss weights $\lambda_{grad}, \lambda_{cont}, \lambda_{style}, \lambda_{hist}, \lambda_{tv}$ \par
 \textbf{Given}: a gradient operator $\triangledown$, a pretrained VGG network $F$ \par
 \textbf{Output}: blending image $I_{B}$ \par
 \For{i $\in$ [1:T]} {
 $I_B = I_Z \odot M + I_T \odot (1-M)$ \par
 $\mathcal{L}_{grad} = GradientLoss(I_B, I_S, I_T, \triangledown)$ by Eq. \ref{Eq:L_grad} \par
 $\mathcal{L}_{cont} = ContentLoss(I_Z, M, I_S, F)$ by Eq. \ref{Eq:L_cont_first} \par
 $\mathcal{L}_{style} = StyleLoss(I_B, I_T, F)$ by Eq. \ref{Eq:L_style_first} \par
 $\mathcal{L}_{hist} = HistogramLoss(I_B, I_T, F)$ by Eq. \ref{Eq:L_hist} \par
 $\mathcal{L}_{tv} = TVLoss(I_B)$ by Eq. \ref{Eq:L_tv} \par
 $L_{total} = 
  \lambda_{grad}*L_{grad}
 +\lambda_{cont}*L_{cont}
 +\lambda_{style}*L_{style}
 +\lambda_{hist}*L_{hist}
 +\lambda_{tv}*L_{tv}$ \par
 $I_Z \leftarrow L$-$BFGS\_Solver(L_{total}, I_Z)$
 }
 $I_B = I_Z \odot M + I_T \odot (1-M)$
 \caption{First Stage - Seamless Blending}
\end{algorithm}

\begin{figure*}[!t]
\centering
\includegraphics[width=\textwidth]{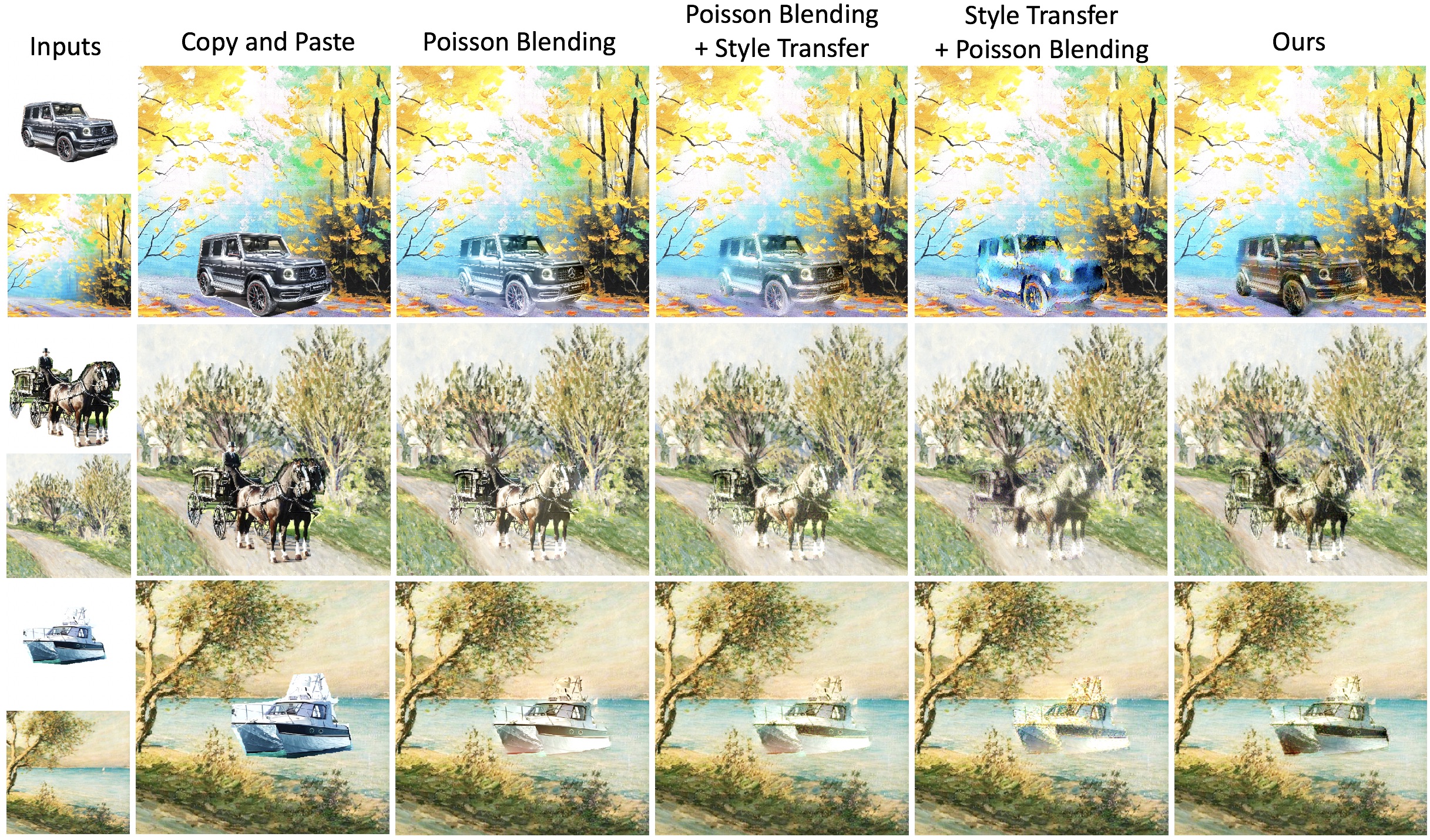}
\caption{This figure shows the comparison between strong baseline approaches and ours on the paintings. \textbf{Poisson Blending} refers to Poisson Image Editing\cite{poisson_editing}. \textbf{Poisson Blending + Style Transfer} refers to first blend the object onto the target image and then run style transfer\cite{style1} on the blending image. \textbf{Style Transfer + Poisson Blending} refers to first run style transfer\cite{style1} on the source image and then blend it onto the target image. }
\label{fig:paint_comp}
\vspace{-10 pt}
\end{figure*}

\begin{algorithm}
\SetAlgoLined
 \textbf{Input}: blending image $I_B$, target image $I_T$\par
 max iteration $T$, loss weights $\lambda_{cont}, \lambda_{style}, \lambda_{hist}, \lambda_{tv}$ \par
 \textbf{Given}: a pretrained VGG network $F$ \par
 \textbf{Output}: refined blending image $I_{BR}$ \par
  $I_{BR} = copy(I_B)$ \par
 \For{i $\in$ [1:T]} {
 $\mathcal{L}_{cont} = ContentLoss(I_{BR}, I_B, F)$ by Eq. \ref{Eq:L_cont_second} \par
 $\mathcal{L}_{style} = StyleLoss(I_{BR}, I_T, F)$ by Eq. \ref{Eq:L_style_second} \par
 $\mathcal{L}_{hist} = HistogramLoss(I_{BR}, I_T, F)$ by Eq. \ref{Eq:L_hist} \par
 $\mathcal{L}_{tv} = TVLoss(I_{BR})$ by Eq. \ref{Eq:L_tv} \par
 $L_{total} = 
  \lambda_{grad}*L_{grad}
 +\lambda_{cont}*L_{cont}
 +\lambda_{style}*L_{style}
 +\lambda_{hist}*L_{hist}
 +\lambda_{tv}*L_{tv}$ \par
 $I_{BR} \leftarrow L$-$BFGS\_Solver(L_{total}, I_{BR})$
 }
 \caption{Second Stage - Style Refinement}
\end{algorithm}

We use VGG layers $conv_{1\_2}$, $conv_{2\_2}$, $conv_{3\_3}$, $conv_{4\_3}$ to compute style loss and $conv_{2\_2}$ to compute content loss. We set maximum iteration to be 1,000 in both stages, and optimize the loss with L-BFGS solver.  The runtime on a 512 $\times$ 512 image takes about 5 minutes on a single NVIDIA GTX 1080Ti. We set $\lambda_{blend} = 10e5, \lambda_{cont} = 1, \lambda_{style} = 10e5, \lambda_{hist} = 1, \lambda_{tv} = 10e-6$ in the second stage, and set $\lambda_{cont} = 1, \lambda_{style} = 10e7, \lambda_{hist} = 1, \lambda_{tv} = 10e-6$ in the second stage. More carefully tuning might give a better combination of hyper-parameters, but this is not our focus.

\section{Experimental Results}

We conduct ablation, comparison and user studies to show the robustness of our method and results. In our ablation study, we show the importance of each loss function and our second stage style refinement. As shown in Fig.(\ref{fig:ablation_study}), our full model clearly outperforms other baseline variants. There exists obvious artifacts on the blending boundary without our proposed blending loss. Some visual contents are wiped out without the content loss. The blending region and target background have inconsistent illumination and texture without style loss. Finally, our two-stage algorithm adds more style and texture in the blending image compared to the single-stage baseline. 

\begin{figure*}[!t]
\centering
\includegraphics[width=\textwidth]{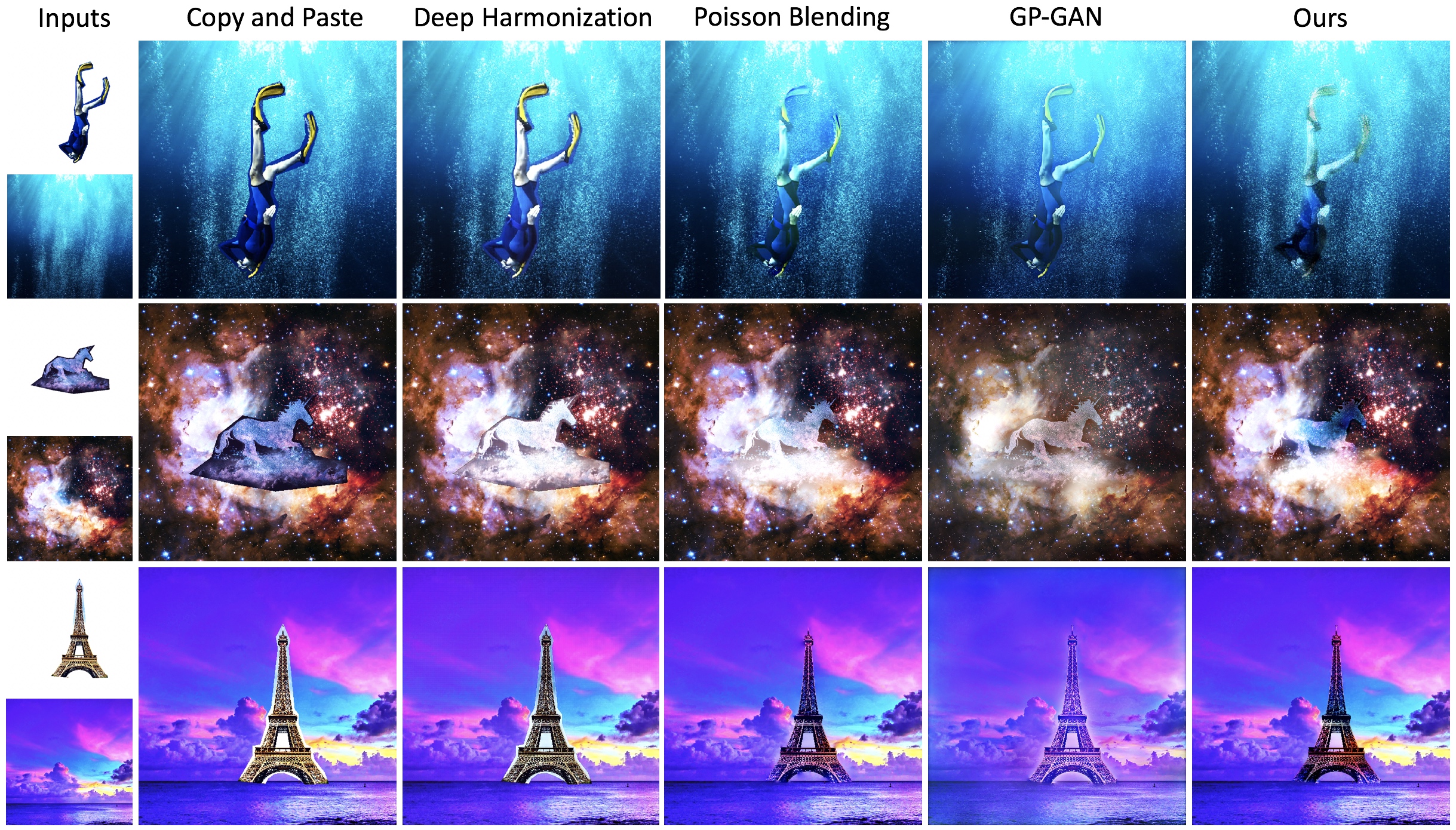}
\caption{This figure shows the comparison between the state-of-the-art image composite approaches and ours on real-world images. 
\textbf{Deep Harmonization} refers to Deep Image Harmonization\cite{harm1}. \textbf{Poisson Blending} refers to Poisson Image Editing\cite{poisson_editing}. \textbf{GP-GAN} refers to Gaussian-Poisson Generative Adversarial Network\cite{gpgan}. }
\label{fig:real_comp}
\vspace{-10 pt}
\end{figure*}

\begin{figure*}[!t]
\centering
\includegraphics[width=\textwidth]{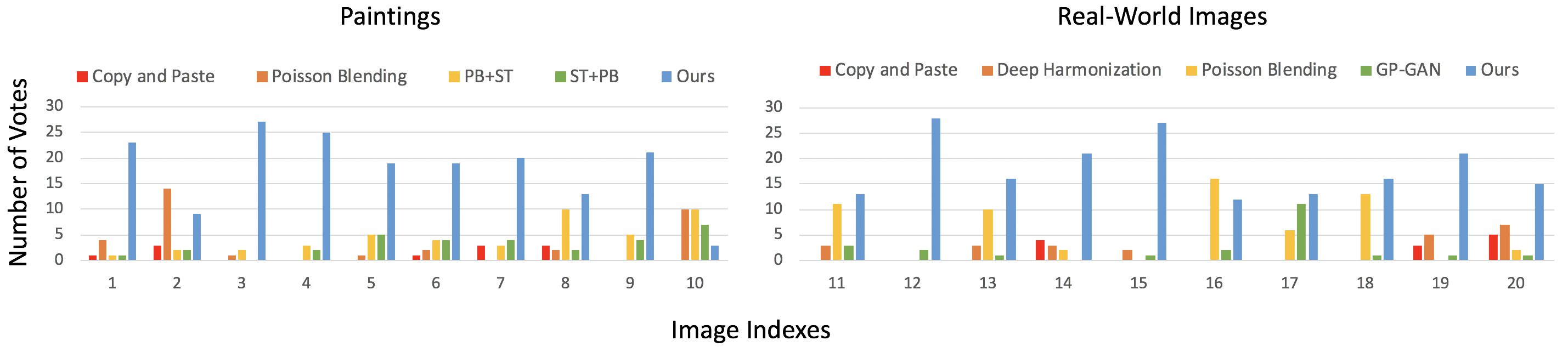}
\vspace{-20 pt}
\caption{This figure shows our user study results. Two histograms on the left and right show the quantitative comparison in paintings and real-world images respectively. "PB+ST" denotes "Poisson Blending + Style Transfer" and "ST+PB" denotes "Style Transfer + Poisson Blending". }
\label{fig:user_study}
\vspace{-10 pt}
\end{figure*}

In our comparison study, we first conduct experiments on paintings as shown in Fig.(\ref{fig:paint_comp}). In this study, we compare with several intuitive and strong baselines. "copy-and-paste" produces results that have obvious artificial boundary. "Poisson Blending"\cite{poisson_editing} is able to produce a smooth blending boundary, but produce inconsistent style and texture between the blending region and the background. "Poisson Blending + Style Transfer"\cite{poisson_editing, style1} produce more consistent style and texture than "Poisson Blending" but produces unpleasant color illumination during Poisson Blending. "Style Transfer + Poisson Blending"\cite{style1, poisson_editing} produces consistent style but lack some of the original source image content. In contrast, our method produces most consistent style and texture while maintaining the source content.

In comparison our study of real-world image, we compare our algorithm with several state-of-the-art image composite algorithms, as shown in Fig.(\ref{fig:real_comp}). "Deep Image Harmonization"\cite{harm6} is a data-driven harmonization algorithm that aims to adjust illumination of composite the region using the learned image prior. As seen, it is able to adjust the illumination of the blending region but is not able to smooth out the boundary, and thus has an artificial border. "Poisson Blending"\cite{poisson_editing} generates seamless boundary, but the background color "blends through" the blending region. "GP-GAN"\cite{gpgan} is a recent proposed blending algorithm that leverages Poisson Blending with Generative Adversarial Network, and is trained in a supervised way. However, this method can hardly generalize to our test cases and produces unrealistic boundary and illumination. Finally, our algorithm produces the best visual results in terms of blending boundary, texture, and color illumination. 

To quantify the performance of our method in comparison to other algorithms, we recruited thirty users for user studies using 20 sets of images, 10 real-world and 10 painting style target images. Each user is asked to pick one composite image they think is the most realistic from five different images generated by five different algorithms. From the histograms (\ref{fig:user_study}), we see that the images that have the highest votes are almost all images generated by our method. Specifically, nine out of ten images our method generated using real-world target images received the highest votes from the users, and eight out of ten images our method generated using painting-style target images received the highest votes from the users. The results indicate that our method is more preferable than other methods 80 $\sim$ 90\% of the time.

\vspace{-5 pt}
\section{Conclusion}
\vspace{-5 pt}

In this paper, we propose a novel gradient blending loss and a two-stage algorithm that generates blending image on-the-fly with a L-BFGS solver. Our algorithm does not rely on any training data and thus can generalize to any real-world images or paintings. Through user study, our algorithm is proven to outperform strong baselines and state-of-the-arts approaches. We believe that our work could be applied in common image editing tasks and opens new possibility for users to easily compose artistic works.

{\small
\bibliographystyle{ieee}
\bibliography{egbib}
}

\end{document}